\documentclass{article} 
\usepackage[T1]{fontenc}
\usepackage{microtype}
\usepackage{graphicx}
\usepackage[table]{xcolor}
\usepackage{subfigure}
\usepackage{booktabs}
\usepackage{amsmath}
\usepackage{amssymb}
\usepackage{multirow}
\usepackage{url}
\usepackage{enumitem}
\usepackage{float}
\usepackage{placeins}
\usepackage{iclr2027_conference,times}

\makeatletter
\renewcommand{\subsection}{\@startsection{subsection}{2}{\z@}{-1.8ex plus
    -0.5ex minus -.2ex}{0.3ex plus .15ex}{\normalsize\sc\raggedright}}
\makeatother

\usepackage{amsmath,amsfonts,bm}

\def\eqref#1{equation~\ref{#1}}

\def\1{\bm{1}}

\DeclareMathAlphabet{\mathsfit}{\encodingdefault}{\sfdefault}{m}{sl}
\SetMathAlphabet{\mathsfit}{bold}{\encodingdefault}{\sfdefault}{bx}{n}

\usepackage[pdfversion=1.7]{hyperref}
\hypersetup{hidelinks,
  pdftitle={CrossPortrait: Cross-Species Portrait Animation via Adversarial Motion Alignment},
  pdfauthor={Gezhong Pan, Zhe Li, Yunfan Tao, Yongheng Zhao, Shiqi Zhou, Juyong Zhang, Xiaosong Jia, Yanyong Zhang, Shuquan Ye, Shaofeng Zhang},
  pdfsubject={Cross-species portrait animation}
}

\title{CrossPortrait: Cross-Species Portrait Animation via Adversarial Motion Alignment}

\author{
Gezhong Pan$^{1,2}$\thanks{Equal contribution.}\thanks{Work done during an internship at Huawei.}\quad
Zhe Li$^{3}$\footnotemark[1]\quad
Yunfan Tao$^{1}$\quad
Yongheng Zhao$^{3}$\quad
Shiqi Zhou$^{3}$\\[3pt]
Juyong Zhang$^{1}$\quad
Xiaosong Jia$^{4}$\quad
Yanyong Zhang$^{1}$\quad
Shuquan Ye$^{2}$\quad
Shaofeng Zhang$^{1}$\thanks{Corresponding author.}\\[3pt]
{\small $^1$University of Science and Technology of China}\\
{\small $^2$Shenzhen Loop Area Institute}\\
{\small $^3$Institute of Media Technology and Experience Design, Huawei}\\
{\small $^4$Fudan University}\\[2pt]
{\small Project page: \url{https://pangezhong.github.io/CrossPortrait/}}
}
\date{}

\newcommand{\ours}{\textsc{CrossPortrait}}
\definecolor{bestyellow}{HTML}{F4C542}
\definecolor{secondyellow}{HTML}{FFF2B2}
\newcommand{\best}[1]{\cellcolor{bestyellow}\textbf{#1}}
\newcommand{\second}[1]{\cellcolor{secondyellow}#1}
\newcommand{\methodna}{N/A}
\newcommand{\selfcomp}[1]{\cellcolor{black}\rule{0pt}{1.6ex}}

\definecolor{src_blue}{HTML}{7EA6E0}
\definecolor{tgt_red}{HTML}{FF9999}
\definecolor{out_green}{HTML}{9AC7BF}

\begin{document}

\maketitle

\begin{figure}[H]
    \centering
    \includegraphics[width=0.88\linewidth]{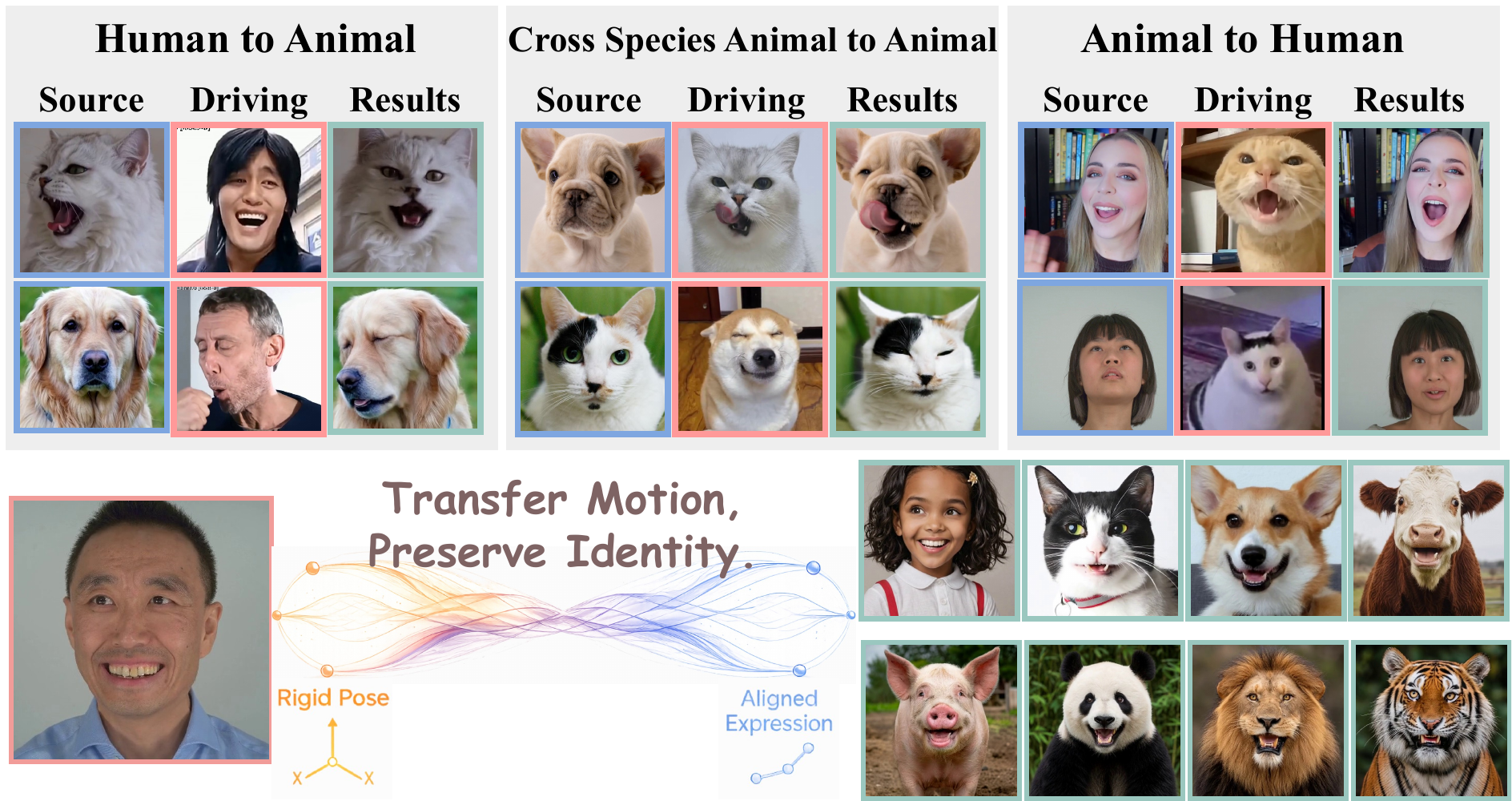}
    \caption{
Cross-species animation by CrossPortrait. 
Our method transfers motion while preserving source identity in human-to-animal, animal-to-animal, and animal-to-human scenarios.
\textcolor{src_blue}{\textbf{Blue}}, \textcolor{tgt_red}{\textbf{red}}, and \textcolor{out_green}{\textbf{green}} borders denote \textcolor{src_blue}{\textbf{source}} portraits, \textcolor{tgt_red}{\textbf{driving}} images, and  \textcolor{out_green}{\textbf{CrossPortrait outputs}}, respectively.
    }
    \label{fig:teaser}
\end{figure}

\begin{abstract}
Cross-species portrait animation, which aims to animate a source portrait of \textit{species A} using a driving video of \textit{species B}, is particularly challenging due to substantial anatomical differences in facial structures across species.
Previous works typically fine-tune human portrait animation models on animal video datasets but often suffer from driving-species leakage.
To address this, we propose \textbf{\ours}, a novel framework for expressive cross-species portrait animation, supporting diverse scenarios including human-to-animal, animal-to-human and cross-species animal-to-animal (e.g. cat-to-dog) animation.
Specifically, we decompose the portrait motion into rigid head poses and expression keypoints. 
To accommodate species-specific facial structures while facilitating cross-species expression alignment, we introduce a species-routed expression encoder that incorporates species-specific LoRAs into a unified expression encoder.
Furthermore, we introduce expression-level adversarial domain adaptation (Ex-ADA) to suppress driving-species-specific information in the expression latents, thereby mitigating species leakage and promoting a shared expression space across species.
Extensive experiments show that our method enables expressive and identity-preserving portrait animation across diverse species and outperforms other state-of-the-art approaches.

\end{abstract}

\section{Introduction}
\label{sec:intro}

Portrait animation synthesizes a video of a source portrait that follows a driving video.
Although AIGC-based methods \citep{zhao2025x,wang2025fantasyportrait,cheng2025wananimate,shi2026dexportrait} have achieved expressive human portrait animation, most of them suffer from generalization to other species, let alone cross-species animation.
In this paper, we focus on \emph{cross-species portrait animation}, where the source and driving input may depict different species, such as humans, cats, or dogs, while the output preserves the source identity, appearance, and species-specific morphology.
We use $X\to Y$ to denote a driving clip from species $X$
applied to a source portrait of species $Y$; for example,
H$\to$A denotes a human driving clip applied to an animal source portrait.
Specifically, our cross-species animation setting mainly includes human-to-animal (H$\to$A), animal-to-human (A$\to$H), and cross-species animal-to-animal (A1$\to$A2) animation, as shown in Fig.~\ref{fig:teaser}.

\begin{figure}[tb!]
   \centering
   \includegraphics[width=\linewidth]{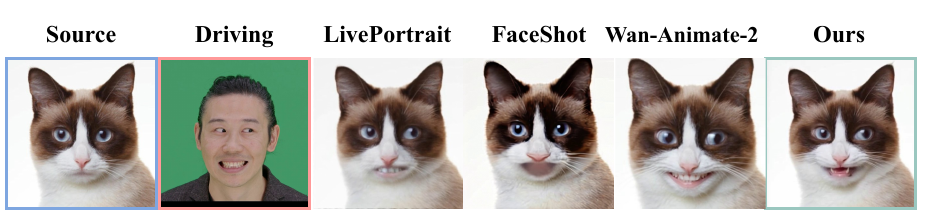}
   \vspace{-15pt}
   \caption{
   Comparison of species leakage with LivePortrait~\citep{guo2024liveportrait},
   FaceShot~\citep{gao2025faceshot}, and Wan-Animate-2~\citep{wang2026wananimate2}.
   These approaches tend to transfer human dental characteristics to the animated cat,
   whereas our method faithfully follows the driving mouth motion while preserving the
   source cat's oral morphology.
   For a fair comparison, we further post-train all applicable baselines on the
    same animal training data.
   }
   \label{fig:motivation_leakage}
\end{figure}

Portrait motion can be decomposed into head pose and facial expression. 
Although head poses across different species can be represented by a shared rigid transformation comprising rotation, translation, and scale (RTS), facial expressions vary substantially across species due to their distinct anatomical structures. 
Therefore, a key challenge lies in learning a shared expression space that aligns semantically corresponding expressions across species, such that the expression of a driving subject can be effectively transferred to a source subject of a different species. 
However, directly encoding the driving image into an expression feature may entangle expression-related information with species-specific cues, leading to species-information leakage.
As illustrated in Fig.~\ref{fig:motivation_leakage}, such leakage may undesirably transfer characteristics of the driving species to the animated source, thereby degenerating source-species fidelity.
Consequently, another key challenge is to effectively suppress species-specific identity cues.

To address these two challenges, we propose an adversarial expression learning framework for cross-species portrait animation. 
For the first challenge, we employ a shared expression encoder to map facial expressions of humans, cats, and dogs into a common representation space, and leverage LivePortrait \citep{guo2024liveportrait} as a training framework to align expression representations across species. 
To accommodate species-specific anatomical variations without compromising the shared representation space, we further introduce a species-routed expression encoder, which incorporates species-specific LoRA \citep{hu2022lora} adapters into a unified driving-side expression encoder. 
The shared backbone facilitates cross-species expression alignment, while the species-specific adapters enhance the representation capacity for capturing fine-grained expression patterns unique to each species. 
For the second challenge, we introduce expression-level adversarial domain adaptation (Ex-ADA) \citep{ganin2016domain}, where a domain classifier is jointly trained to predict the species category from the encoded expression features. 
Through adversarial training with a gradient reversal layer (GRL), the expression encoder is encouraged to suppress species-specific identity cues, yielding more species-agnostic expression representations and thereby mitigating driving-species leakage.
Finally, we reshape the driving expression features into expression tokens and inject them into the video DiT \citep{wan2025wan} via cross-attention, enabling expressive cross-species portrait animation while preserving source-species fidelity.




Overall, our contributions are summarized as follows:
\vspace{-7pt}
\begin{itemize}[leftmargin=*]
    \item We propose \textbf{CrossPortrait}, a unified dual-path framework for cross-species portrait-motion alignment: rigid head motion is represented with dense pose-flow, while facial deformation is represented with shared implicit-keypoint tokens, covering human-to-animal, animal-to-human, and cross-species animal-to-animal animation.
    \item We introduce \textbf{species-routed expression encoder}, which inserts species-specific LoRA adapters into a unified driving-side expression encoder, preserving capacity for anatomical differences while providing a shared expression-conditioning interface.
    \item We introduce \textbf{expression-level adversarial domain adaptation} (Ex-ADA), applied to expression features to suppress driving-species-specific information, mitigate species leakage, and avoid adversarially aligning rigid pose or source morphology.
\end{itemize}

\section{Related Work}
\noindent\textbf{Motion representations for portrait animation.}\quad
Portrait animation methods condition generation on motion representations with varying levels of structure.
Parametric models provide interpretable pose and expression coefficients: PIRenderer~\citep{ren2021pirenderer}, for example, uses coefficients from a 3D morphable model.
Self-supervised approaches instead discover sparse keypoints or regions and convert their motion into dense warps~\citep{siarohin2019first,siarohin2021motion,zhao2022thin}.
FaceVid2Vid~\citep{wang2021facevid2vid} decomposes learned 3D keypoints into canonical geometry, rigid pose, and non-rigid deformation, while LivePortrait~\citep{guo2024liveportrait} extends this interface with stitching and retargeting.
Dense motion can also condition video generators: MOFA-Video~\citep{niu2024mofa} converts sparse controls into motion fields, while FG-Portrait~\citep{xu2026fgportrait} derives 3D motion flow from a parametric head model.
Other methods learn motion conditions directly from images or latent features.
X-Portrait~\citep{xie2024xportrait} extracts hierarchical features from driving frames, whereas HunyuanPortrait and X-NeMo~\citep{xu2025hunyuanportrait,zhao2025x} inject compact implicit motion representations through attention.
DeX-Portrait~\citep{shi2026dexportrait} combines explicit global pose with latent expression, and DPE~\citep{pang2023dpe} studies pose--expression disentanglement.
Wan-Animate~\citep{cheng2025wananimate} uses causal convolutions to compress an implicit face representation to the video-latent frame rate.
Overall, explicit geometry offers a stable, interpretable interface but is tied to predefined anatomy, whereas learned representations are more flexible but not inherently comparable across domains.
We therefore model rigid head motion geometrically and facial deformation with learned implicit-keypoint tokens.
Each keypoint trajectory is refined at the original frame rate before the retained tokens are packed to match the VAE timeline.

\par\smallskip
\noindent\textbf{Portrait animation beyond human faces.}\quad
Most portrait-animation methods study cross-identity transfer within the human-face domain, including X-Portrait~\citep{xie2024xportrait}, EMOPortraits~\citep{drobyshev2024emoportraits}, Follow-Your-Emoji~\citep{ma2024followyouremoji,ma2025followyouremojifaster}, HunyuanPortrait~\citep{xu2025hunyuanportrait}, and X-NeMo~\citep{zhao2025x}.
A smaller body of work broadens the source domain.
Animate-X~\citep{tan2025animatex} animates anthropomorphic characters from pose guidance, FaceShot~\citep{gao2025faceshot} retargets facial landmarks to diverse characters, FantasyPortrait~\citep{wang2025fantasyportrait} reports animal-source examples, and LivePortrait provides an animal model.
These works mainly show that human-derived motion can animate broader source appearances, rather than extracting motion from non-human drivers.
Animal-driven portrait animation remains less explored.
FaceShot includes several non-human driving examples, although its CharacBench evaluation uses human drivers.
AnimateZoo~\citep{xu2024animatezoo} studies cross-species motion transfer, but focuses on full-body actions and mismatched skeletons rather than portrait-scale head pose and facial deformation.
Our work instead considers human, cat, and dog sources and drivers, covering human-to-animal, animal-to-human, same-species animal, and cross-species animal-to-animal animation within one shared generator.

\par\smallskip
\noindent\textbf{Adversarial domain alignment and domain-specific routing.}\quad
Domain-adversarial learning reduces domain-predictive information by training a domain classifier through a gradient reversal layer~\citep{ganin2016domain}.
Related approaches separate shared and domain-private features~\citep{bousmalis2016domain}, condition alignment on task predictions~\citep{long2018conditional}, or use multiple discriminators for finer-grained alignment~\citep{pei2018multi}.
Residual adapters provide domain-specific capacity around a shared model~\citep{rebuffi2017learning}, while low-rank adaptation offers parameter-efficient routed updates~\citep{hu2022lora}.
We adopt this dual-path interface for cross-species motion conditioning; related video-generation work explores similar alignment and decoupling choices~\citep{zhang2025videorepa,zheng2026vpt}.
At inference, the driving-species label selects a species-routed LoRA adapter in the expression encoder to capture species-specific appearance and anatomy. During Stage 1 training, Ex-ADA is applied to both source- and driving-frame expression features from same-species reconstruction pairs; at inference, only the driving-side expression features are used for motion conditioning. The rigid-pose pathway, source-reference pathway, and shared video generator remain species-agnostic and are not adversarially aligned.

\section{Method}

\subsection{Problem Formulation and Overview}

\begin{figure}[tb!]
    \centering
    \makebox[\linewidth][c]{%
        \includegraphics[width=1\linewidth]{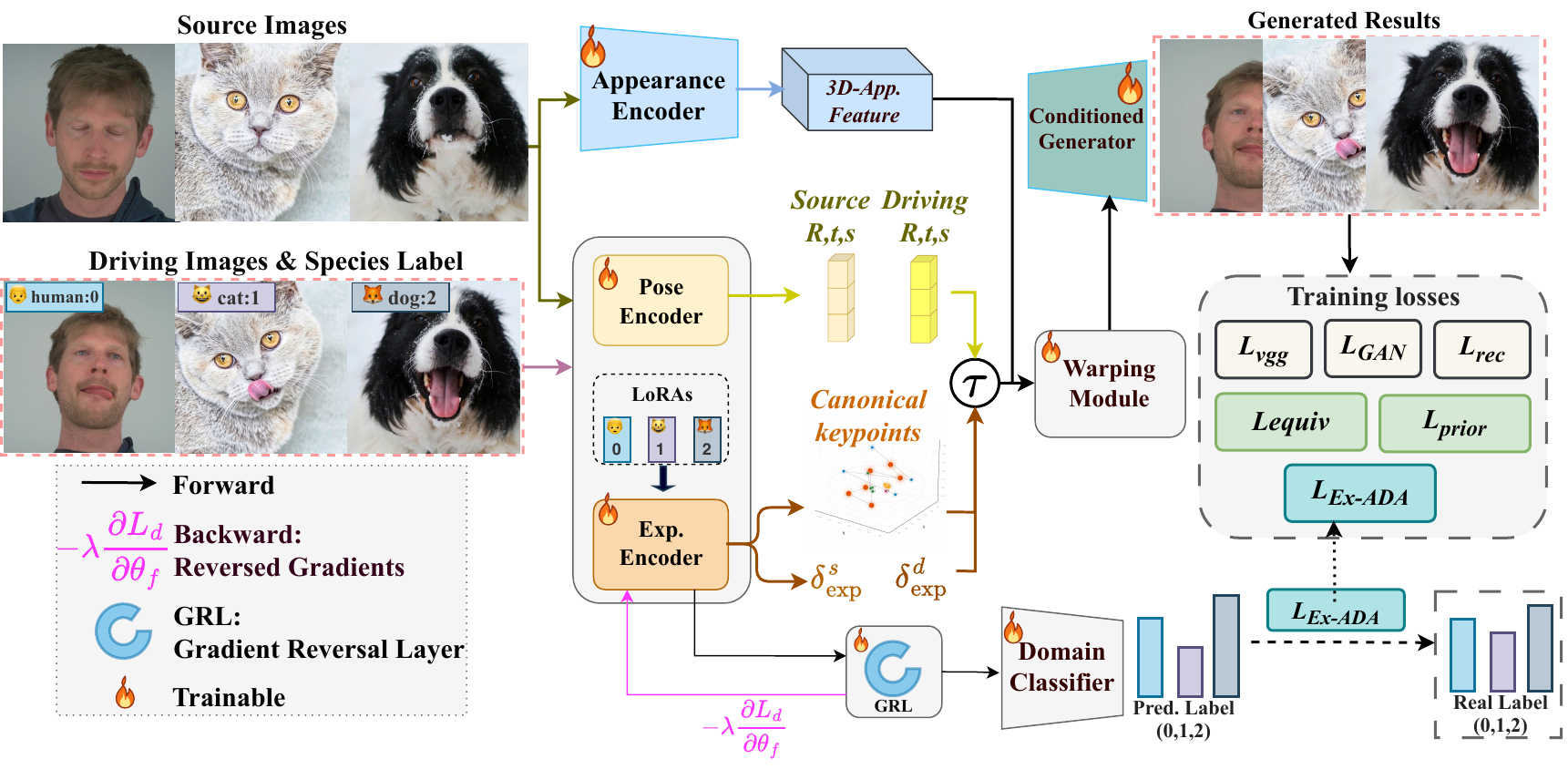}%
    }
    \caption{
    \textbf{Stage 1: Learning cross-species motion alignment.} We align expression representations across species while suppressing
species-specific identity cues through the species-routed expression encoder and expression-level adversarial domain adaptation (Ex-ADA).
    }
    \label{fig:stage1_pipeline}
\end{figure}

\begin{figure}[tb!]
    \centering
    \makebox[\linewidth][c]{%
        \includegraphics[width=1.1\linewidth]{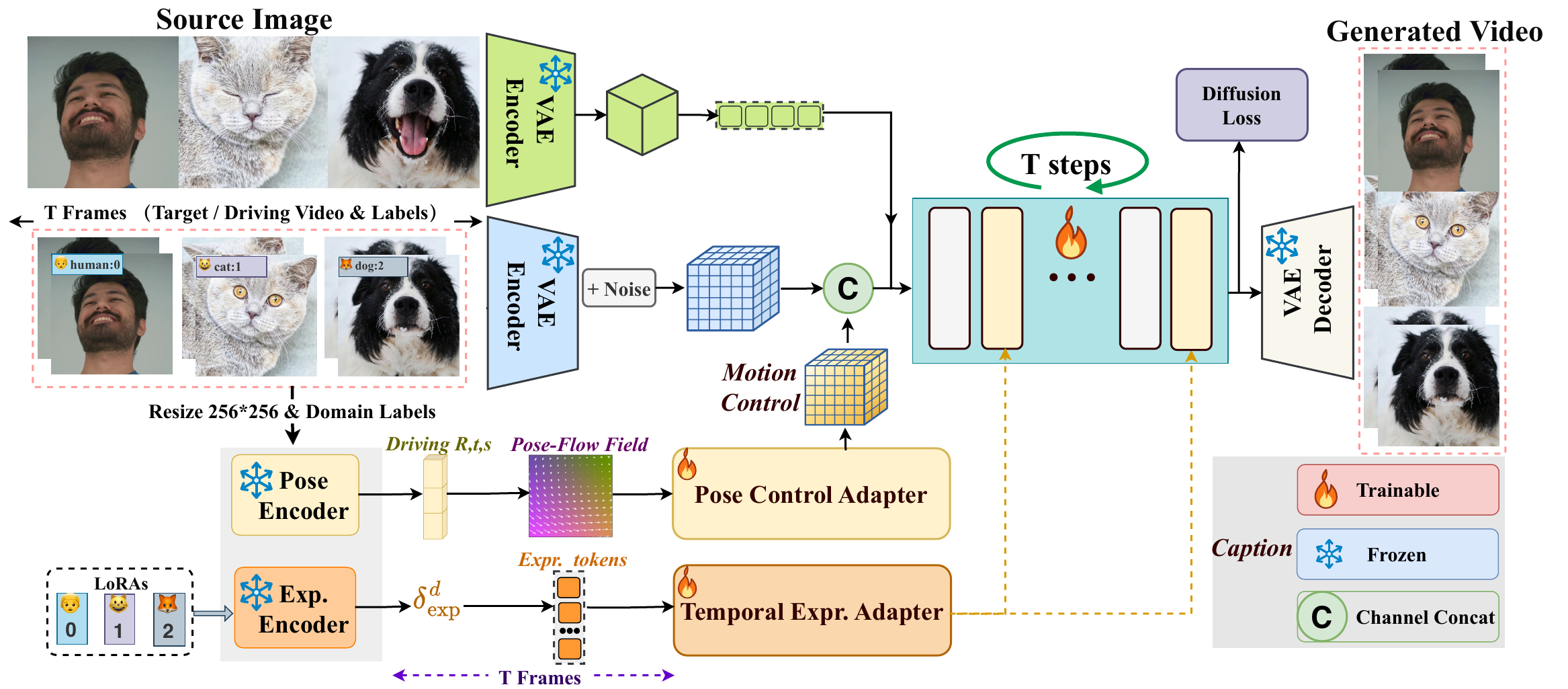}%
    }
    \caption{
    \textbf{Stage 2: DiT-based cross-species portrait animation.} 
    We inject the pose and expression features extracted from pretrained pose and expression encoders into the video DiT for expressive cross-species animation.
    }
    \label{fig:stage2_pipeline}
\end{figure}

Given a source portrait $I_s$ and a driving video
$V_d=(I_d^t)_{t=1}^{T}$, possibly from different species, we synthesize
\begin{equation}
\hat{V}
=
G\left(I_s, M(V_d)\right),
\end{equation}
where $\hat{V}$ preserves the source identity, appearance, and \textbf{species-specific morphology} while following the driving motion, $M(\cdot)$ denotes the motion extractor, and $G(\cdot)$ is the generator.
Unlike previous human portrait animation methods, our goal is to enable \textbf{cross-species} animation, although the training datasets contain only same-species pairs including human, cat, and dog videos.
As illustrated in Fig.~\ref{fig:stage1_pipeline} and Fig.~\ref{fig:stage2_pipeline}, our method contains two stages:

\vspace{-4pt}
\begin{enumerate}[leftmargin=*]
    \item \textbf{Learning cross-species motion alignment}.
    We employ LivePortrait \citep{guo2024liveportrait} for motion alignment. Specifically, rigid head motion is represented by an affine transformation, while facial expression is encoded as shared implicit keypoints.
    First, we augment the shared expression backbone with LoRA components used by the species-routed expression encoder, enhancing the encoder's capacity to model anatomically distinct expression patterns while preserving a common representation across species.
    We then introduce expression-level adversarial domain adaptation (Ex-ADA) to suppress driving-species-specific information in the expression latents, thereby mitigating species leakage and facilitating cross-species expression alignment.

    \item \textbf{DiT-based cross-species portrait animation.}
    Given the pretrained pose and expression encoders from Stage 1, we inject head pose as a motion flow and expression features as cross-attention tokens into the DiT-based video generation backbone. 
    Benefiting from the well-aligned expression encoders with suppressed species leakage, the model achieves faithful cross-species animation despite being trained solely on same-species data pairs.
    
\end{enumerate}

\subsection{Learning Cross-species Motion Alignment}

Cross-species motion transfer requires a common expression representation
that accommodates anatomical differences while suppressing driving-species
cues. 
We learn this representation within the LivePortrait \citep{guo2024liveportrait} reconstruction
framework using same-species source--driving pairs from human, cat, and
dog videos.


\vspace{-8pt}
\paragraph{Motion representations.}
We decompose portrait motion into rigid head pose and non-rigid facial
expression. For an input portrait $I$ with species label
$c \in \{\mathrm{human}, \mathrm{cat}, \mathrm{dog}\}$, we estimate
\begin{equation}
\begin{aligned}
(R,t,s) &= E_{\mathrm{pose}}(I),
\quad
\delta_{\mathrm{exp}} = H_{\mathrm{exp}}\!\left(E_{\mathrm{exp}}(I;c)\right).
\end{aligned}
\end{equation}
Here, $R$, $\mathbf{t}$, and $s$ denote rotation, translation, and scale,
respectively. The expression encoder produces an intermediate feature
$F_{\mathrm{exp}} = E_{\mathrm{exp}}(I;c) \in \mathbb{R}^{768}$,
which is mapped by the shared prediction head $H_{\mathrm{exp}}$ to
$K=21$ implicit 3D keypoint offsets
$\boldsymbol{\delta}_{\mathrm{exp}} \in \mathbb{R}^{K \times 3}$.
During Stage~1, these motion components are combined with canonical
keypoints to drive the shared warping and reconstruction modules, while
appearance is provided by the source portrait.

\vspace{-8pt}
\paragraph{Species-routed expression encoder.}
A shared expression backbone provides a common representation space, but
must accommodate anatomically distinct expression patterns. We therefore augment selected backbone layers with LoRA components used by the
species-routed expression encoder (Hu et al., 2022). The backbone and
expression prediction head are shared across species, while these components
provide additional capacity for interpreting subtle species-dependent facial
expressions. As illustrated on the right of Fig.~\ref{fig:ablation_qualitative}, the LoRA
components used by the species-routed expression encoder enable more accurate
expression reenactment.


\vspace{-8pt}
\paragraph{Expression-level adversarial domain adaptation (Ex-ADA).}
Species-specific adaptation alone does not prevent expression features from retaining anatomical or appearance cues that can leak into the animated source. 
We therefore introduce expression-level adversarial domain adaptation (Ex-ADA). 
A three-class (human, cat, and dog) species classifier $D$ takes the expression feature as input through a gradient reversal layer (GRL). 
Given a same-species source--driving pair $(I_s,I_d)$ with species label $c$, we define the ADA loss as
\begin{equation}
\mathcal{L}_{\mathrm{ADA}}
=
\frac{1}{2}
\sum_{i \in \{s,d\}}
\operatorname{CE}
\left(
D\left(\operatorname{GRL}(F_{\mathrm{exp}}^{i})\right),
c
\right),
\label{eq:exada}
\end{equation}
where $F_{\mathrm{exp}}^{i}=E_{\mathrm{exp}}(I_{i};c)$ and
$\operatorname{CE}$ denotes cross-entropy.
The GRL acts as an identity mapping during the forward pass, while reversing and scaling the gradient during backpropagation.
The ADA loss trains the classifier to predict species labels, while the reversed gradients encourage the expression encoder to suppress species-specific information in its features.

Ex-ADA complements the reconstruction objective: reconstruction and motion priors encourage the representation to retain expression-related information, while adversarial training discourages reliance on species cues. 
Together, they promote a shared expression interface for cross-species transfer without paired cross-species supervision.
As shown on the left of Fig.~\ref{fig:ablation_qualitative}, Ex-ADA enables expression transfer while better preserving the source animal's species-specific oral morphology.

\subsection{DiT-Based Cross-Species Portrait Animation}

We freeze the motion encoders learned in Stage~1 and condition a shared
video DiT on driving pose and expression, while the reference pathway
provides source appearance and morphology.

\vspace{-8pt}
\paragraph{Pose condition.}
Following DeX-Portrait~\citep{shi2026dexportrait}, we convert the
predicted rotation, translation, and scale into dense pose-flow fields.
These fields are temporally packed and projected to match the VAE
latents, then concatenated with the noisy latents to provide spatial
motion guidance.

\vspace{-8pt}
\paragraph{Expression condition.}
For each driving frame $t$, we convert the predicted expression offsets
$\delta_{\mathrm{exp}}^t$ into keypoint-wise expression tokens.
Specifically, for the $k$-th keypoint, we first obtain an initial token
$z_{t,k}$ and refine its temporal trajectory:
\begin{equation}
\begin{aligned}
z_{t,k} &= \phi(\delta_{\mathrm{exp},k}^{t}) + p_k,
\qquad &
\widetilde{z}_{t,k}
&= z_{t,k} + \mathcal{G}_{\mathrm{causal}}(z_{\leq t,k}),
\quad k=1,\ldots,K .
\end{aligned}
\end{equation}
where $\phi$ is a shared MLP that projects each 3D keypoint offset into a
token embedding, $p_k$ is a learned keypoint embedding, and
$\mathcal{G}_{\mathrm{causal}}$ applies left-padded causal convolutions
independently to each keypoint trajectory.
Refinement
operates at the original frame rate, and its final projection is
zero-initialized to preserve the initial frame-wise representation.

To match the VAE timeline, we prepend three copies of the first token
frame and pack every four consecutive frames into $4K$ expression tokens
per latent time step. After projection to the DiT hidden dimension,
these tokens serve as keys and values in expression cross-attention
adapters, with video tokens as queries. The adapters are inserted every
two transformer blocks and use zero-initialized output projections.
This supplies temporally refined expression guidance through
the shared cross-species expression representations learned in Stage~1.

\subsection{Training}

In Stage 1, pose and expression encoders~\citep{liu2022convnet} are initialized from pretrained human weights of LivePortrait~\citep{guo2024liveportrait} and jointly optimized with reconstruction, motion prior, and ADA losses.
The warping and reconstruction modules are shared across species.
In Stage 2, pose and expression encoders are frozen.
We train the DiT denoising backbone, motion projections, causal expression refiner, and shared expression cross-attention adapters on human, cat, and dog videos.
Neither stage introduces generator-side species routing to encourage learning facial structures from the source image.
Complete objectives and optimization settings are provided in Appendix~\ref{app:method_details}.
\section{Experiments}
\label{sec:experiments}
We evaluate motion fidelity and identity preservation across three transfer settings: human-to-animal (H$\to$A), animal-to-human (A$\to$H), and animal-to-animal (A$\to$A). For A$\to$A, we use balanced cat--dog transfer as the primary trained-species setting and additionally assess generalization to unseen source species qualitatively. We further conduct controlled studies with a joint No-Alignment baseline and
ablations of Ex-ADA, the species-routed expression encoder, and temporal stabilization.

\subsection{Experimental Setup}
\label{sec:exp_setup}

\vspace{-4pt}
\paragraph{Data and splits.}
Both stages use the same training data, comprising public human portrait datasets and internally collected cat and dog videos. Stage 1 uses two sampled frames per video for reconstruction training, while Stage 2 uses continuous 81-frame clips.
All evaluation identities and driving clips are held out from training.
The animal self-reenactment benchmark contains 50 cat and 50 dog videos; the remaining evaluation covers H$\to$A, A$\to$H, cross- and same-species A$\to$A, and unseen-source transfer with trained-species drivers.
Dataset details, preprocessing, resolutions, and split construction are provided in Appendix~\ref{app:data_protocol}.

\vspace{-8pt}
\paragraph{Compared methods.}
We group the compared methods by their roles and supported settings. LivePortrait-Animal~\citep{guo2024liveportrait} serves as our principal external animal-trained baseline for H$\to$A, A$\to$H, and the applicable A$\to$A settings. FaceShot~\citep{gao2025faceshot}, Follow-Your-Emoji~\citep{ma2024followyouremoji}, FantasyPortrait~\citep{wang2025fantasyportrait}, and Wan-Animate-2~\citep{wang2026wananimate2} provide additional comparisons in the cross-species settings they support. FantasyPortrait's released pipeline relies on human-face detection/alignment and pretrained facial-motion features, so it does not officially support animal videos as drivers. Its animal examples use human drivers for animal portraits; we therefore retain it only for H$\to$A and mark animal self-reenactment as N/A. Human-only methods are evaluated only in the H$\to$H sanity check because they
do not support the cross-species settings. Appendix~\ref{app:h2h_sanity} shows that \ours{} achieves the best or tied-best H$\to$H scores on all metrics except APD, on which it ranks second. The \textbf{No-Alignment Baseline} uses the same human--cat--dog data and generator backbone, providing a matched-data joint baseline for the cross-species motion interface; component ablations are reported separately below. For each supported setting, all methods are evaluated using identical source--driving pairs and their official preprocessing.
Each applicable baseline is post-trained on the same animal training data
using its official pipeline. Methods with partial applicability are evaluated only on the settings they support, and unsupported method--direction pairs are marked N/A. Protocol-specific applicability is summarized in Appendix Table~\ref{tab:baseline_applicability}.

\vspace{-8pt}
\paragraph{Evaluation metrics.}
For paired animal self-reenactment, we use the first frame of each target video as the source image and the remaining frames as ground-truth targets.
We report peak signal-to-noise ratio (PSNR), structural similarity (SSIM)~\citep{wang2004image}, learned perceptual image patch similarity (LPIPS)~\citep{zhang2018unreasonable}, and $L_1$ error.
For A$\to$H, we additionally report CSIM between the generated human face and the source portrait to measure source identity preservation; CSIM is not used to measure motion fidelity.
For same-species animal cross-identity transfer, we report pairwise preferences in Appendix Table~\ref{tab:same_species_results}.
AED and APD are reported only for human frames.
Detailed metric definitions are provided in Appendix~\ref{app:metric_details}.

\vspace{-8pt}
\paragraph{Pairwise user study.}
Because the evaluated cross-species and animal cross-identity settings lack
paired ground truth and a validated animal-motion evaluator, H$\to$A, A$\to$H,
and cross- and same-species A$\to$A trials use anonymized pairwise preferences.
Participants view the source portrait, a synchronized driving clip, and two anonymized outputs, with the output order randomized and counterbalanced.
They independently judge four criteria: Expr. measures expression fidelity and temporal alignment; Pose measures head-pose agreement; the ID criterion measures preservation of source identity and morphology, including driving-species leakage; and Overall measures motion, identity and morphology preservation, temporal coherence, and visual quality.
The study enrolled 50 participants, each completing 10 comparison trials and five ablation trials, yielding 500 comparison and 250 ablation trials in total.
A$\to$H pairs are balanced between cat and dog drivers. For cross-species A$\to$A, we use equal numbers of cat$\to$dog and dog$\to$cat pairs, and every regime is balanced across the motion strata defined in Appendix~\ref{app:data_protocol}.
Every question includes a tie option, with ties assigned a preference score of 0.5. We report Method over \ours{} preference rates. Lower values indicate better performance by \ours{}. Failures of applicable official pipelines on the fixed evaluation pairs are counted as losses, while unsupported method--direction pairs are marked N/A and excluded from the analysis.

\subsection{Cross-Species Results}

\begin{figure}[tb!]
    \centering
    \includegraphics[width=\linewidth]{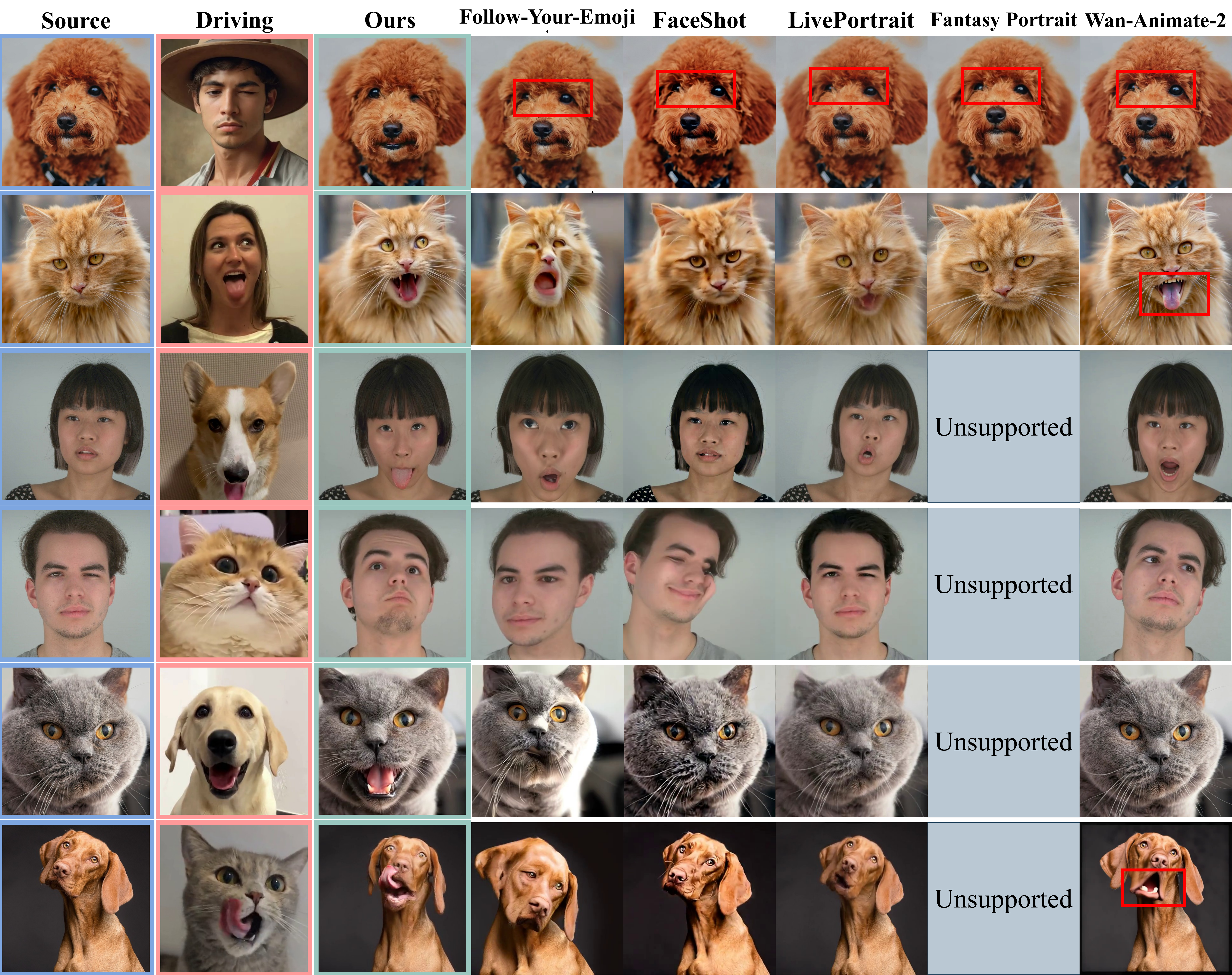}
    \caption{
    Qualitative comparison of \ours{} with competing methods.
    The top, middle, and bottom rows show human-to-animal, animal-to-human, and
    cross-species cat$\leftrightarrow$dog animation, respectively.
    \textcolor{src_blue}{\textbf{Blue}},
    \textcolor{tgt_red}{\textbf{red}}, and
    \textcolor{out_green}{\textbf{green}} borders indicate
    \textcolor{src_blue}{\textbf{source}} portraits,
    \textcolor{tgt_red}{\textbf{driving}} images, and
    \textcolor{out_green}{\textbf{CrossPortrait outputs}}, respectively.
    Red boxes mark visible artifacts, and gray cells indicate unsupported cases.
    }
    \label{fig:qualitative}
\end{figure}

\vspace{-4pt}
\paragraph{Quantitative comparison.}
Tables~\ref{tab:animal_h2a_results} and~\ref{tab:animal_driven_results}
report reconstruction, CSIM, and pairwise preference results across animal
self-reenactment and cross-species transfer. Figure~\ref{fig:qualitative}
shows fixed-pair examples, and Appendix Table~\ref{tab:human_results} reports
the H$\to$H sanity check. \ours{} achieves the best self-reenactment metrics
and highest A$\to$H CSIM, while every valid baseline-over-\ours{} preference
rate remains below 50\% across cross-species settings.

\providecommand{\selfslash}{%
    \raisebox{0.08ex}{\scalebox{0.9}{/}}%
}

\begin{table}[tb!]
\centering
\caption{
Animal self-reenactment reconstruction and H$\to$A pairwise preference evaluation.
Self-reenactment uses 50 cat and 50 dog videos.
Each preference value denotes the rate at which users prefer the corresponding method over \ours{}.
}
\label{tab:animal_h2a_results}
\resizebox{\linewidth}{!}{
\begin{tabular}{lcccccccc}
\toprule
Method
& \multicolumn{4}{c}{Animal Self-Reenactment}
& \multicolumn{4}{c}{H$\to$A User Preference: Method over \ours{}} \\
\cmidrule(lr){2-5}
\cmidrule(lr){6-9}
& PSNR $\uparrow$
& SSIM $\uparrow$
& LPIPS $\downarrow$
& $L_1$ $\downarrow$
& Expr. (\%) $\downarrow$
& Pose (\%) $\downarrow$
& ID (\%) $\downarrow$
& Overall (\%) $\downarrow$ \\
\midrule

Follow-Your-Emoji
& 21.38
& 0.7639
& 0.1730
& 0.05030
& 0.0
& 0.0
& 0.0
& 0.0 \\

FaceShot
& 16.17
& 0.6019
& 0.3822
& 0.1012
& 3.2
& 1.6
& 12.9
& 3.2 \\

LivePortrait-Animal
& 24.37
& 0.7720
& 0.1438
& 0.04010
& 21.2
& 21.2
& 34.6
& 23.1 \\

FantasyPortrait
& \methodna
& \methodna
& \methodna
& \methodna
& 0.0
& 0.0
& 12.9
& 0.0 \\

Wan-Animate-2
& 23.54
& \second{0.7834}
& 0.1955
& 0.0590
& 30.0
& 22.5
& 34.2
& 19.8
\\

No-Alignment Baseline
& \second{24.59}
& 0.7743
& \second{0.1400}
& \second{0.04001}
& 23.6
& 26.9
& 14.9
& 26.2 \\

\textbf{\ours{}}
& \best{25.06}
& \best{0.7893}
& \best{0.1360}
& \best{0.03970}
& \selfslash
& \selfslash
& \selfslash
& \selfslash \\

\bottomrule
\end{tabular}
}
\end{table}

\begin{table}[tb!]
\centering

\caption{
A$\to$H and cross-species A$\to$A (A1$\to$A2)  evaluation.
A$\to$H uses balanced cat and dog driving clips.
Each value denotes the preference rate for the corresponding method over \ours{}; lower values indicate better performance by \ours{}.
}

\label{tab:animal_driven_results}
\resizebox{\linewidth}{!}{
\begin{tabular}{lccccccccc}
\toprule
Method
& \multicolumn{5}{c}{A$\to$H (Balanced cat/dog drivers)}
& \multicolumn{4}{c}{Cross-species A$\to$A (A1$\to$A2) } \\
\cmidrule(lr){2-6}
\cmidrule(lr){7-10}
& CSIM $\uparrow$
& Expr. (\%) $\downarrow$
& Pose (\%) $\downarrow$
& ID (\%) $\downarrow$
& Overall (\%) $\downarrow$
& Expr. (\%) $\downarrow$
& Pose (\%) $\downarrow$
& ID (\%) $\downarrow$
& Overall (\%) $\downarrow$ \\
\midrule

FaceShot
& \second{0.8657}
& 0.0
& 0.0
& 6.2
& 0.0
& 0.0
& 0.0
& 0.0
& 0.0 \\

Follow-Your-Emoji
& 0.8445
& 0.0
& 0.0
& 0.0
& 0.0
& 0.0
& 0.0
& 0.0
& 0.0 \\

LivePortrait-Animal
& 0.8625
& 0.0
& 0.0
& 0.0
& 0.0
& 0.0
& 0.0
& 12.5
& 0.0 \\

Wan-Animate-2
& 0.8629
& 0.4
& 8.8
& 4.2
& 7.9
& 6.2
& 8.2
& 0.0
& 3.8 \\

No-Alignment Baseline
& 0.8451
& 18.2
& 8.2
& 0.0
& 12.2
& 23.9
& 17.1
& 10.9
& 20.6 \\

\textbf{\ours{}}
& \best{0.8825}
& \selfslash
& \selfslash
& \selfslash
& \selfslash
& \selfslash
& \selfslash
& \selfslash
& \selfslash \\

\bottomrule
\end{tabular}
}
\end{table}

\vspace{-8pt}

\paragraph{Qualitative comparison.}
Figures~\ref{fig:teaser} and~\ref{fig:qualitative} present bidirectional
cross-species transfer, including unseen source portraits driven by trained
species. Baselines show motion errors, geometric distortion, and
species-specific feature leakage, while \ours{} better preserves source
morphology and follows the driving motion.

\subsection{Ablation Studies}
\label{sec:ablation}

All ablation variants use the same data, source--driving pairs, and generator backbone,
with the corresponding user-study results summarized in
Table~\ref{tab:ablation_results}.

\begin{figure}[tb!]
    \centering
    \begin{minipage}[t]{0.55\linewidth}
        \centering
        \includegraphics[width=\linewidth]
        {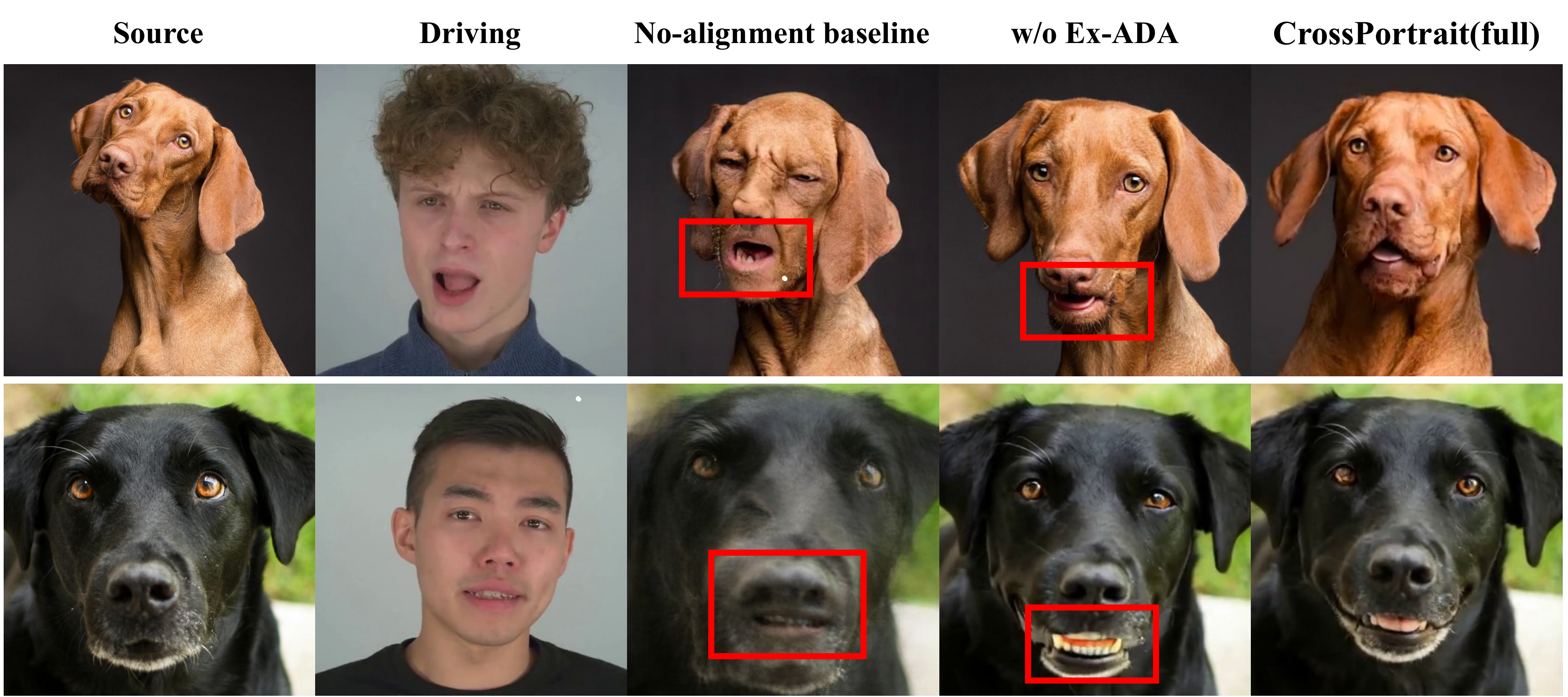}
    \end{minipage}
    \hfill
    \begin{minipage}[t]{0.44\linewidth}
        \centering
        \includegraphics[width=\linewidth]
        {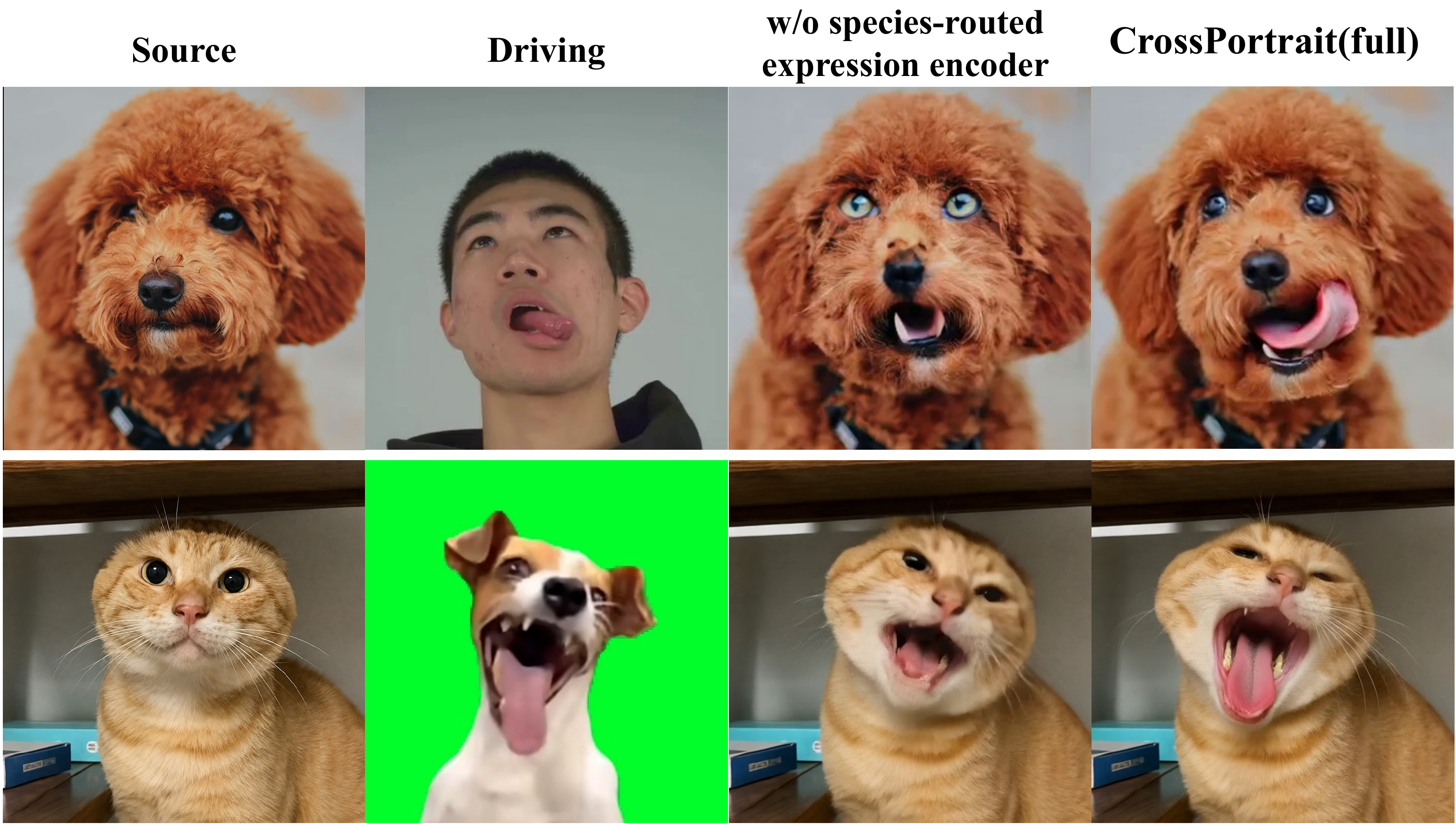}
    \end{minipage}
    \caption{
    Qualitative H$\to$A ablations.
    Left: No-Alignment and w/o expression-level ADA.
    Right: w/o species-routed expression encoder.
    }
    \label{fig:ablation_qualitative}
\end{figure}

\noindent\textbf{No-Alignment Baseline.}
The No-Alignment variant uses a single shared ConvNeXt encoder to process both
pose-flow and expression-token motion. It removes the LoRA components used by
the species-routed expression encoder and does not apply expression-level
adversarial domain adaptation (Ex-ADA).
This configuration substantially weakens cross-species motion transfer and can
yield unstable mouth geometry
(Figure~\ref{fig:ablation_qualitative}, left).

\noindent\textbf{w/o Ex-ADA.}
Removing only expression-level ADA leaves more driving-species information in
the expression features, increasing species-specific feature leakage in
H$\to$A transfer
(Figure~\ref{fig:ablation_qualitative}, left).
ADA suppresses species-predictive information at the expression level without
applying adversarial alignment to the pose-flow or source-reference pathways.

\noindent\textbf{w/o species-routed expression encoder.}
Removing the species-routed expression encoder leaves a shared expression encoder to model species-dependent anatomical variations, weakening fine-grained motions such as tongue protrusion and mouth opening.
The species-routed expression encoder restores the ability to model variations while retaining a shared expression backbone (Figure~\ref{fig:ablation_qualitative}, right).

\noindent\textbf{w/o temporal stabilization.}
Replacing causal temporal refinement with non-causal processing allows future
driving states to influence earlier expression tokens, leading to less temporally
consistent mouth and tongue motion. This results in lower Expression and Overall
preference rates, and can also slightly degrade perceived pose quality during
joint generation.

\begin{table}[tb!]
    \centering
\caption{
Ablation user study on bidirectional human--animal animation.
Each value denotes the preference rate for the corresponding method over \ours{}; lower values indicate better performance by \ours{}.
}
\label{tab:ablation_results}
\resizebox{\linewidth}{!}{
\begin{tabular}{lcccc}
\toprule
Variant
& Expr.  (\%) $\downarrow$
& Pose  (\%) $\downarrow$
& ID  (\%) $\downarrow$
& Overall  (\%) $\downarrow$ \\
\midrule

(A) No-alignment baseline
& 6.4
& 15.4
& 11.5
& 14.1 \\

(B) w/o Ex-ADA
& 23.1
& 26.9
& 23.1
& 19.2 \\

(C) w/o species-routed expression encoder
& 21.2
& 30.3
& 33.3
& 27.9 \\

(D) w/o temporal stabilization.
& 30.0
& 48.0
& 44.0
& 32.0 \\

\bottomrule
\end{tabular}
}
\end{table}

\FloatBarrier

\section{Limitations}

Our training data cover human, cat, and dog drivers. Generalization to unseen species is therefore evaluated only qualitatively on the source side, using drivers from the trained species, while unseen-species driving and arbitrary facial topologies remain unexplored.
Moreover, cross-species animation lacks paired ground truth and validated cross-species motion metrics. As a result, preference-based evaluation and identity metrics cannot fully characterize motion accuracy or biological plausibility.

\section{Conclusion}

We presented \ours{}, a unified framework for cross-species portrait animation across human, cat, and dog domains.
Rather than enforcing domain invariance over all motion information, \ours{} uses complementary pathways for explicit geometric motion and learned expression conditioning, retains species-specific modeling capacity through the species-routed expression encoder, and applies alignment only to expression features to suppress species-predictive information.
A lightweight causal temporal module further stabilizes expression-motion tokens before video generation.
Our results show that this design enables a single shared generator to support diverse source--driving directions while improving motion transfer and source preservation over applicable baselines.
Ablation studies further indicate the importance of balancing shared motion representations with species-specific modeling capacity for cross-species animation.
While the model generalizes qualitatively to unseen source species when driven by trained species, unseen driver species remain unexplored.
Extending the framework to a broader range of species and less structured motion domains is a promising direction for future work.

\section*{Reproducibility Statement}
We describe the model architecture, training objectives, optimization settings,
data construction, evaluation splits, metric definitions, and user-study protocol
in the main paper and the supplementary appendices. All evaluation source identities
and driving clips are held out from training, and the implementation details
needed to reproduce the reported experiments are provided in the supplementary
material. Qualitative video results are available on the project page:
\url{https://pangezhong.github.io/CrossPortrait/}.

\section*{AI Assistance}
We used large language models to assist with language editing and manuscript organization, including improving grammar, clarity, and presentation. All AI-assisted text was reviewed and revised by the authors, who take full responsibility for the final manuscript, claims, and references.

\bibliography{crossportrait}
\bibliographystyle{iclr2027_conference}

\clearpage
\appendix

\begin{center}
    {\Large\bfseries Supplementary Material\par}
\end{center}
\vspace{1em}

\newcommand{\appmainline}[3]{%
    \noindent\textbf{#1\quad #2}\dotfill #3\par}
\newcommand{\appsubline}[3]{%
    \noindent\hspace{1.6em}#1\quad #2\dotfill #3\par}

\appmainline{A}{\hyperref[app:method_details]{Method and Training Details}}
            {\pageref{app:method_details}}
\appsubline{A.1}{\hyperref[app:method_pose]{Pose Representation and Pose-Flow Conditioning}}
            {\pageref{app:method_pose}}
\appsubline{A.2}{\hyperref[app:method_sampling]{Adversarial Schedule and Dataset Sampling}}
            {\pageref{app:method_sampling}}
\appsubline{A.3}{\hyperref[app:method_stage1]{Stage-1 Reconstruction Objective}}
            {\pageref{app:method_stage1}}
\appmainline{B}{\hyperref[app:metric_details]{Evaluation Metric Details}}
            {\pageref{app:metric_details}}
\appsubline{B.1}{\hyperref[app:metric_notation]{Notation and Temporal Aggregation}}
            {\pageref{app:metric_notation}}
\appsubline{B.2}{\hyperref[app:metric_aed]{Average Expression Distance}}
            {\pageref{app:metric_aed}}
\appsubline{B.3}{\hyperref[app:metric_apd]{Average Pose Distance}}
            {\pageref{app:metric_apd}}
\appsubline{B.4}{\hyperref[app:metric_csim]{Cosine Similarity}}
            {\pageref{app:metric_csim}}
\appmainline{C}{\hyperref[app:h2h_sanity]{Human-to-Human Sanity Check}}
            {\pageref{app:h2h_sanity}}
\appsubline{C.1}{\hyperref[app:h2h_setup]{Evaluation Setup}}
            {\pageref{app:h2h_setup}}
\appmainline{D}{\hyperref[app:baseline_applicability]{Baseline Applicability}}
            {\pageref{app:baseline_applicability}}
\appsubline{D.1}{\hyperref[app:baseline_matrix]{Applicability Matrix}}
            {\pageref{app:baseline_matrix}}
\appmainline{E}{\hyperref[app:implementation_details]{Implementation Details}}
            {\pageref{app:implementation_details}}
\appsubline{E.1}{\hyperref[app:impl_stage1_arch]{Stage-1 Network Architecture}}
            {\pageref{app:impl_stage1_arch}}
\appsubline{E.2}{\hyperref[app:impl_stage1_opt]{Stage-1 Optimization}}
            {\pageref{app:impl_stage1_opt}}
\appsubline{E.3}{\hyperref[app:impl_stage2_cond]{Stage-2 Motion Conditioning}}
            {\pageref{app:impl_stage2_cond}}
\appsubline{E.4}{\hyperref[app:impl_stage2_opt]{Stage-2 Optimization}}
            {\pageref{app:impl_stage2_opt}}
\appmainline{F}{\hyperref[app:same_species]{Same-Species Animal Evaluation}}
            {\pageref{app:same_species}}
\appsubline{F.1}{\hyperref[app:same_species_setup]{Evaluation Setting}}
            {\pageref{app:same_species_setup}}
\appmainline{G}{\hyperref[app:data_protocol]{Data and Evaluation Protocol}}
            {\pageref{app:data_protocol}}
\appsubline{G.1}{\hyperref[app:data_training]{Training Data}}
            {\pageref{app:data_training}}
\appsubline{G.2}{\hyperref[app:data_splits]{Evaluation Splits}}
            {\pageref{app:data_splits}}
\appsubline{G.3}{\hyperref[app:data_user_study]{Pairwise User-Study Protocol}}
            {\pageref{app:data_user_study}}
\clearpage

\section{Method and Training Details}
\label{app:method_details}

\subsection{Pose Representation and Pose-Flow Conditioning}
\label{app:method_pose}

\begin{figure}[tb!]
    \centering
    \includegraphics[width=\linewidth]{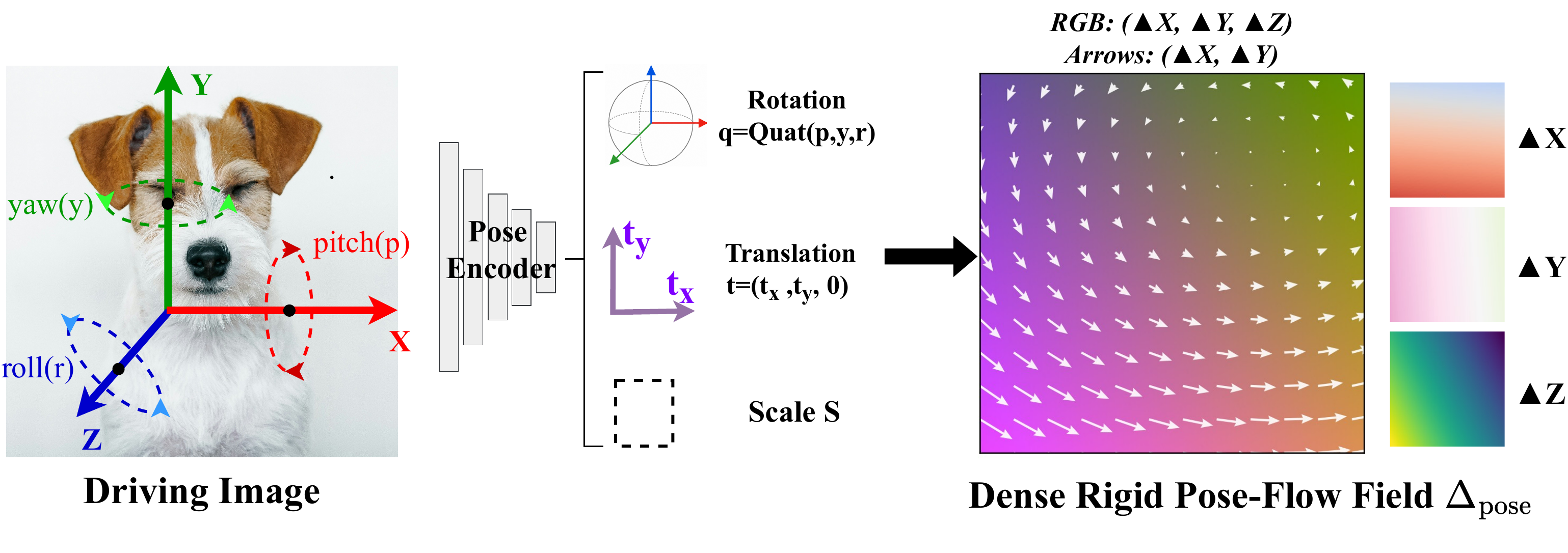}
    \caption{
    Pose branch for dense rigid pose-flow.
    PoseNet predicts pitch, yaw, roll, translation, and scale; these parameters
    are converted into the dense three-channel displacement field
    $\Delta_{\mathrm{pose}}$.
    }
    \label{fig:pose_branch}
\end{figure}

\paragraph{Pose parameterization.}
PoseNet predicts pitch, yaw, and roll, together with translation and scale:
\begin{equation}
p,y,r=H_{\mathrm{rot}}(F_{\mathrm{pose}}),\qquad
t=H_t(F_{\mathrm{pose}}),\qquad
s=H_s(F_{\mathrm{pose}}).
\end{equation}
Each rotation angle is obtained by taking a soft expectation over
$B_{\mathrm{bin}}$ bins,
$\theta=\sum_i\mathrm{softmax}(l)_i b_i$,
where $l$ denotes the angle logits and $b_i$ the corresponding bin centers.
The resulting Euler angles are converted to a quaternion
$q=\mathrm{Quat}(p,y,r)$.
For translation, we use $t=(t_x,t_y,0)$ in normalized crop coordinates.
These pose parameters are further converted into a dense rigid pose-flow field,
as illustrated in Figure~\ref{fig:pose_branch}.
For video conditioning, four consecutive pose-flow fields are channel-packed,
resulting in 12 channels, which are projected by a $1\times1\times1$ Conv3D
to the 48-channel VAE latent-control dimension.

\subsection{Adversarial Schedule and Dataset Sampling}
\label{app:method_sampling}

After a 3,000-step warm-up, the GRL multiplier used by Ex-ADA follows
\begin{equation}
\lambda(p)=\frac{2}{1+\exp(-\gamma p)}-1,
\label{eq:grl_schedule}
\end{equation}
where $p$ denotes the normalized training progress and $\gamma=10$.
The sampling weights of the human, cat, and dog datasets have total masses
of 6.7, 7.5, and 7.5, respectively, corresponding to approximately
30.9\%, 34.6\%, and 34.6\% of the training mixture.
Training batches are shuffled from this mixture rather than explicitly
constructed to be class-balanced.

\subsection{Stage-1 Reconstruction Objective}
\label{app:method_stage1}

The full reconstruction-stage loss is
\begin{equation}
\begin{aligned}
\mathcal{L}_{\mathrm{stage1}}={}&
\lambda_{\mathrm{rec}}\mathcal{L}_{\mathrm{rec}}
+\lambda_{\mathrm{per}}\mathcal{L}_{\mathrm{per}}
+\lambda_{\mathrm{equiv}}\mathcal{L}_{\mathrm{equiv}}
+\lambda_{\mathrm{kp}}\mathcal{L}_{\mathrm{kp}}^{\mathrm{prior}}\\
&+\lambda_{\mathrm{def}}\mathcal{L}_{\mathrm{def}}^{\mathrm{prior}}
+\lambda_{\mathrm{gan}}\mathcal{L}_{\mathrm{gan}}
+\lambda_{\mathrm{Ex-ADA}}\mathcal{L}_{\mathrm{Ex-ADA}}.
\end{aligned}
\end{equation}
All terms operate on Stage-1 source--driving reconstruction pairs.
Stage 1 contains no temporal operation, and causal processing is used only in the Stage-2 expression refiner.
The reconstruction and perceptual terms preserve image fidelity, the
equivariance term regularizes motion consistency, and the keypoint and
deformation priors stabilize the predicted geometry. The global GAN term
improves local realism, while Ex-ADA penalizes species-predictive information
in the expression representation through the gradient-reversal pathway.

\section{Evaluation Metric Details}
\label{app:metric_details}

\subsection{Notation and Temporal Aggregation}
\label{app:metric_notation}

Here, $\hat{I}^t$ and $I_d^t$ denote the generated and driving frames at
time $t$, respectively, and $T$ denotes the number of evaluated frames.

\subsection{Average Expression Distance (AED)}
\label{app:metric_aed}

AED measures the accuracy of facial expression transfer using parameters
extracted by SMIRK~\citep{retsinas2024smirk}. Following the evaluation
protocol of prior work, we construct the expression vector by concatenating
the facial-expression, eyelid, and jaw-pose parameters predicted by SMIRK:
\begin{equation}
E_{\mathrm{SMIRK}}(I)
=
\left[
\mathbf{e}(I),\,
\mathbf{l}(I),\,
\mathbf{j}(I)
\right],
\end{equation}
where $\mathbf{e}$, $\mathbf{l}$, and $\mathbf{j}$ denote the facial-expression,
eyelid, and jaw-pose parameters, respectively. AED is computed as the mean
absolute difference between the expression vectors of the generated and
driving frames:
\begin{equation}
\mathrm{AED}
=
\frac{1}{T D_E}
\sum_{t=1}^{T}
\left\|
E_{\mathrm{SMIRK}}(\hat{I}^t)
-
E_{\mathrm{SMIRK}}(I_d^t)
\right\|_1 ,
\end{equation}
where $D_E$ is the dimensionality of the expression vector.
Lower AED indicates better agreement with the driving expression.

\subsection{Average Pose Distance (APD)}
\label{app:metric_apd}

APD measures head-pose accuracy using the pose parameters predicted by
SMIRK. Let $P_{\mathrm{SMIRK}}(I)$ denote the SMIRK head-pose parameter
vector. APD is computed as
\begin{equation}
\mathrm{APD}
=
\frac{1}{T D_P}
\sum_{t=1}^{T}
\left\|
P_{\mathrm{SMIRK}}(\hat{I}^t)
-
P_{\mathrm{SMIRK}}(I_d^t)
\right\|_1 ,
\end{equation}
where $D_P$ is the dimensionality of the pose vector.
Lower APD indicates better agreement with the driving head pose.

\subsection{Cosine Similarity (CSIM)}
\label{app:metric_csim}

Identity preservation is evaluated using cosine similarity between
ArcFace~\citep{deng2019arcface} embeddings.
For cross-identity reenactment, CSIM is computed between each generated
frame and the source portrait:
\begin{equation}
\mathrm{CSIM} =
\frac{1}{T}\sum_{t=1}^{T}
\mathrm{cos}
\left(
A_{\mathrm{ArcFace}}(\hat{I}^t),
A_{\mathrm{ArcFace}}(I_s)
\right),
\end{equation}
where $A_{\mathrm{ArcFace}}(\cdot)$ denotes the pretrained ArcFace embedding
extractor, and $\mathrm{cos}(\cdot,\cdot)$ denotes cosine similarity.
Higher CSIM indicates better preservation of the source identity.

\section{Human-to-Human Sanity Check}
\label{app:h2h_sanity}

\subsection{Evaluation Setup}
\label{app:h2h_setup}

\begin{figure}[tb!]
\centering
\includegraphics[width=\linewidth]{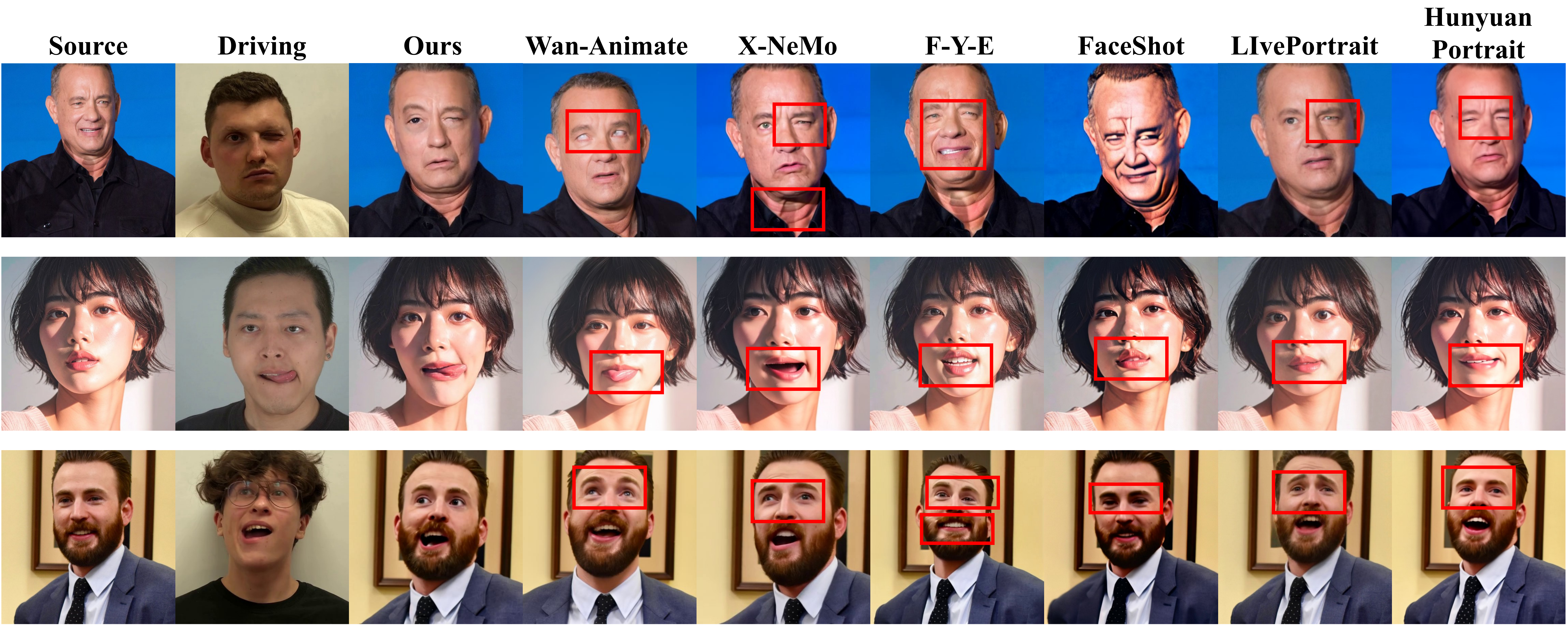}
\caption{
Qualitative H$\to$H cross-identity comparison on fixed source--driving pairs.
Columns show the source portrait, driving frame, and outputs from \ours{},
Wan-Animate, X-NeMo, Follow-Your-Emoji, FaceShot, LivePortrait-Human,
and HunyuanPortrait.
Red boxes highlight local artifacts in the eye or mouth regions.
}
\label{fig:h2h_qualitative}
\end{figure}

Human-to-human animation is not the primary focus of this work.
We nevertheless include an H$\to$H evaluation as a sanity check to examine whether multi-species training and the proposed motion-conditioning design maintain standard human portrait-animation performance.
We evaluate on 100 held-out human videos from Pixabay~\citep{pixabay} and Pexels~\citep{pexels}, using the same preprocessing protocol as in the animal evaluation.
For self-reenactment, the first frame serves as the source portrait and the remaining frames provide pixel-level ground truth.
For cross-identity reenactment, the source and driving videos contain different identities.
We compare with Follow-Your-Emoji, LivePortrait-Human, FaceShot,
HunyuanPortrait~\citep{xu2025hunyuanportrait},
X-NeMo~\citep{zhao2025x}, and
Wan-Animate~\citep{cheng2025wananimate}.
For self-reenactment, we report PSNR, SSIM, LPIPS, $L_1$, AED, APD, and CSIM.
For cross-identity reenactment, where pixel-level ground truth is unavailable, we report the unpaired human-face metrics AED, APD, and CSIM.
Table~\ref{tab:human_results} and Figure~\ref{fig:h2h_qualitative} report the
quantitative and qualitative comparisons, respectively.

\begin{table}[tb!]
\centering
\caption{
Human-to-human sanity check.
Self-reenactment is evaluated with pixel-level ground truth, while
cross-identity reenactment uses unpaired human motion and identity metrics.
}
\label{tab:human_results}
\resizebox{\linewidth}{!}{
\begin{tabular}{lcccccccc}
\toprule
Method
& \multicolumn{5}{c}{Self-Reenactment}
& \multicolumn{3}{c}{Cross-Identity Reenactment} \\
\cmidrule(lr){2-6}
\cmidrule(lr){7-9}
& PSNR $\uparrow$
& SSIM $\uparrow$
& LPIPS $\downarrow$
& $L_1$ $\downarrow$
& CSIM $\uparrow$
& AED $\downarrow$
& APD $\downarrow$
& CSIM $\uparrow$ \\
\midrule

Follow-Your-Emoji
& 22.20
& 0.7650
& 0.1204
& 0.0428
& 0.8130
& 0.3039
& 0.0540
& 0.4282 \\

LivePortrait-Human
& 20.83
& 0.7419
& 0.1443
& 0.0518
& \second{0.8750}
& 0.3328
& 0.2615
& 0.5528 \\

FaceShot
& 17.99
& 0.6836
& 0.2382
& 0.0696
& 0.7078
& 0.3868
& 0.3104
& 0.2971 \\

HunyuanPortrait
& 21.46
& 0.7584
& 0.1328
& 0.0459
& 0.8589
& 0.2270
& 0.1182
& \second{0.7208} \\

X-NeMo
& 21.12
& 0.7501
& 0.1397
& 0.0486
& 0.8463
& \second{0.2250}
& 0.0200
& 0.6650 \\

Wan-Animate
& \second{24.05}
& \second{0.8065}
& \second{0.1055}
& \second{0.0316}
& 0.8407
& 0.2402
& \best{0.0190}
& 0.6056 \\

\textbf{\ours{}}
& \best{24.20}
& \best{0.8156}
& \best{0.1016}
& \best{0.0305}
& \best{0.8765}
& \best{0.2241}
& \second{0.0191}
& \best{0.7308} \\

\bottomrule
\end{tabular}
}
\end{table}

\section{Baseline Applicability}
\label{app:baseline_applicability}

\subsection{Applicability Matrix}
\label{app:baseline_matrix}

Table~\ref{tab:baseline_applicability} distinguishes source-side animal
generalization from support for animal driving.
Cross-species A$\to$A is evaluated quantitatively with balanced
cat$\leftrightarrow$dog pairs; additional qualitative examples use unseen
species only as sources.

\begin{table}[tb!]
\centering
\caption{
Applicability of compared methods under our evaluation protocol.
}
\label{tab:baseline_applicability}
{\small
\setlength{\tabcolsep}{7pt}
\renewcommand{\arraystretch}{1.08}
\begin{tabular}{lccccc}
\toprule
Method
& H$\to$H
& \shortstack{Same-species\\A$\to$A}
& H$\to$A
& A$\to$H
& \shortstack{Cross-species\\A$\to$A} \\
\midrule

\multicolumn{6}{l}{\textit{Cross-domain and animal-capable baselines}} \\

Follow-Your-Emoji
& Yes
& Partial
& Yes
& Partial
& Partial \\

LivePortrait-Animal
& \methodna
& Yes
& Yes
& Yes
& Partial \\

FaceShot
& Yes
& Yes
& Yes
& Yes
& Yes \\

FantasyPortrait
& Yes
& \methodna
& Yes
& \methodna
& \methodna \\

Wan-Animate-2
& Yes
& Yes
& Yes
& Yes
& Yes \\

\addlinespace
\multicolumn{6}{l}{\textit{Human-only baselines}} \\

LivePortrait-Human
& Yes
& \methodna
& \methodna
& \methodna
& \methodna \\

HunyuanPortrait
& Yes
& \methodna
& \methodna
& \methodna
& \methodna \\

X-NeMo
& Yes
& \methodna
& \methodna
& \methodna
& \methodna \\

Wan-Animate
& Yes
& \methodna
& \methodna
& \methodna
& \methodna \\

\addlinespace
\multicolumn{6}{l}{\textit{Matched-data comparison}} \\

No-Alignment Baseline
& Yes
& Yes
& Yes
& Yes
& Yes \\

\textbf{\ours{}}
& Yes
& Yes
& Yes
& Yes
& Yes \\

\bottomrule
\end{tabular}
}
\smallskip

\begin{minipage}{0.96\linewidth}
\footnotesize
\textit{Protocol notes.}
``Yes'' denotes an applicable setting covered by the evaluation protocol,
while ``Partial'' denotes cases in which valid outputs are available for only
a subset of source--driving pairs.
N/A denotes an officially unsupported setting and is excluded from the
corresponding evaluation.
``Failed'' denotes an applicable setting that was attempted but did not produce
a valid output; such failures are retained in the evaluation rather than
discarded.
LivePortrait-Human and LivePortrait-Animal are separate released pipelines;
the latter supports animation with animal data and is evaluated on
animal-driven settings where valid outputs can be obtained.
FantasyPortrait is used as a zero-shot H$\to$A baseline in our cross-species
evaluation and does not officially support animal driving.
Wan-Animate and Wan-Animate-2 are treated as separate baselines: the former is
used in the H$\to$H sanity check, while the latter is evaluated in
cross-species and animal-driven settings.
\end{minipage}
\end{table}

\section{Implementation Details}
\label{app:implementation_details}

\subsection{Stage-1 Network Architecture}
\label{app:impl_stage1_arch}

PoseNet and ExpNet use separate ConvNeXt backbones initialized from the
corresponding weights of the human LivePortrait checkpoint and are jointly
trained under the Stage-1 objectives.
Both networks take $256\times256$ portrait crops as input.
PoseNet uses 66-bin heads for pitch, yaw, and roll, with soft expectation
converting the predicted distributions to angles at a $3^{\circ}$ interval
before quaternion conversion.
The expression head predicts 21 shared 3D keypoint offsets.
The species-routed expression encoder is implemented using LoRA residuals
inserted only into the pwconv2 linear layers of the last three blocks of
ConvNeXt stage 3 and all three blocks of stage 4, corresponding to zero-based
stage indices 2 and 3.
Each human, cat, and dog route uses rank 8, $\alpha=8$, and dropout 0.05.
The LoRA up-projection is zero-initialized, such that all routed branches are
initially identical to the shared backbone.

The three-class species discriminator receives the 768-dimensional ExpNet
backbone feature and has linear dimensions
$768\rightarrow256\rightarrow64\rightarrow3$.
The two hidden layers use LeakyReLU with slope 0.2 and dropout 0.3; Gaussian
noise with standard deviation 0.1 is added to discriminator inputs during
training.
The GRL is disabled for the first 3,000 optimization steps and then follows
the schedule in Eq.~(\ref{eq:grl_schedule}) with $\gamma=10$.
Its expression-gradient scale is 0.30.
Ex-ADA is applied only to the expression branch and not to the pose branch.

\subsection{Stage-1 Optimization}
\label{app:impl_stage1_opt}

Stage 1 uses mean squared error (MSE) reconstruction, VGG perceptual,
equivariance, keypoint-prior, deformation-prior, global image generative
adversarial network (GAN), and Ex-ADA losses.
Their respective weights are
0.1, 0.1, 1.0, 1.0, 1.0, 0.5, and 0.3.
We optimize with Adam using $\beta=(0,0.999)$.
The learning rates are $5\times10^{-5}$ for PoseNet and the remaining
generator modules, $10^{-5}$ for the shared ExpNet backbone,
$2\times10^{-5}$ for the expression keypoint heads,
$5\times10^{-5}$ for the species-routed LoRA parameters, and
$10^{-4}$ for the image discriminator.
Training runs for approximately three days on four GPUs using bfloat16 mixed
precision, a per-GPU batch size of 20 without gradient accumulation,
gradient clipping at 1.0, and seed 42.

\subsection{Stage-2 Motion Conditioning}
\label{app:impl_stage2_cond}

Stage 2 uses a Wan2.2-Fun-5B-Control backbone and a frozen
Wan VAE~\citep{wan2025wan}.
Training clips contain 81 frames at $512\times512$ with temporal stride 1;
shorter clips are padded by repeating sampled frames.
Driving frames are resized to $256\times256$ for MotionNet.
The frozen pose branch converts absolute pose into a three-channel dense
pose-flow field at the VAE latent spatial resolution.
Four consecutive fields are channel-packed, with the first field repeated to
align the first latent time step, and a $1\times1\times1$ Conv3D projects the
resulting 12 channels to the 48-channel VAE latent control dimension.

Each of the 21 three-dimensional expression offsets is mapped by a
$3\rightarrow64\rightarrow512$ multilayer perceptron with Gaussian error linear
unit (GELU) activations and added to a learned keypoint embedding.
The token-wise causal residual refiner applies non-affine LayerNorm followed by
left-padded temporal convolutions $512\rightarrow1024\rightarrow512$, both with
kernel size 3, and a sigmoid linear unit (SiLU) activation between them.
The final convolution is zero-initialized, and the encoder output is added
residually to preserve temporal length and provide an identity initialization.
After prepending three copies of the first frame, every four frames are packed
into 84 tokens and linearly projected to the DiT hidden dimension.
Expression cross-attention adapters are inserted every two transformer blocks,
and their output projections are zero-initialized.
Expression conditioning is independently dropped for each training sample with
probability 0.2.

\subsection{Stage-2 Optimization}
\label{app:impl_stage2_opt}

The VAE, PoseNet, ExpNet, and Stage-1 species-routed LoRA parameters are frozen
in Stage 2.
We train the full DiT denoising backbone, the
$3\rightarrow64\rightarrow512$ expression projection and positional embedding,
the pose Conv3D projection, the causal expression refiner, and the expression
cross-attention adapters.
The flow-matching target is the difference between sampled Gaussian noise and
the clean video latent, optimized with the configured sigma-weighted MSE
objective.
We use AdamW with $\beta=(0.9,0.999)$, weight decay 0.03, and
$\epsilon=10^{-10}$.
The main DiT learning rate is $2\times10^{-5}$, the patch-embedding rate is
$10^{-5}$, and the learning rate for motion projections and expression
cross-attention adapters is $10^{-4}$.
The learning-rate scheduler is constant; the configuration specifies 500
warmup steps per process and the distributed launcher passes
$500\times4=2{,}000$ warmup steps to the scheduler.
Training uses four GPUs, bfloat16 precision, gradient checkpointing, a per-GPU
batch size of 1 without accumulation, gradient clipping at 0.05, and seed 42.
Checkpoints and validation samples are produced every 1,500 and 750
optimization steps, respectively.

\section{Same-Species Animal Evaluation}
\label{app:same_species}

\subsection{Evaluation Setting}
\label{app:same_species_setup}

Table~\ref{tab:same_species_results} reports the supplementary same-species
animal cross-identity user study.
Because no broadly accepted species-aware motion evaluator or paired
cross-identity ground truth is available, animal-to-animal cross-identity
evaluation uses pairwise human preference rather than a human-face metric.

\begin{table}[tb!]
\centering
\caption{
Same-species animal cross-identity user study.
Entries are pairwise preference rates for \ours{} over each evaluated baseline,
with ties counted as half a vote. The preference columns are percentages (\%):
Expr. measures expression fidelity and temporal alignment; Pose measures
head-pose agreement; ID measures source identity and morphology preservation,
including driving-species leakage; and Overall measures motion, source
preservation, temporal coherence, and visual quality.
}
\label{tab:same_species_results}
\begin{tabular}{lccccc}
\toprule
Method
& Valid $n$
& Expr. (\%) $\uparrow$
& Pose (\%) $\uparrow$
& ID (\%) $\uparrow$
& Overall (\%) $\uparrow$ \\
\midrule

FaceShot
& 50
& 100
& 100
& 100
& 100 \\

Follow-Your-Emoji
& 50
& 100
& 100
& 100
& 100 \\

LivePortrait-Animal
& 50
& 100
& 100
& 91
& 100 \\

Wan-Animate-2
& 50
& 100
& 98
& 100
& 100 \\

\bottomrule
\end{tabular}
\end{table}

\begin{figure}[tb!]
    \centering
    \includegraphics[width=\linewidth]{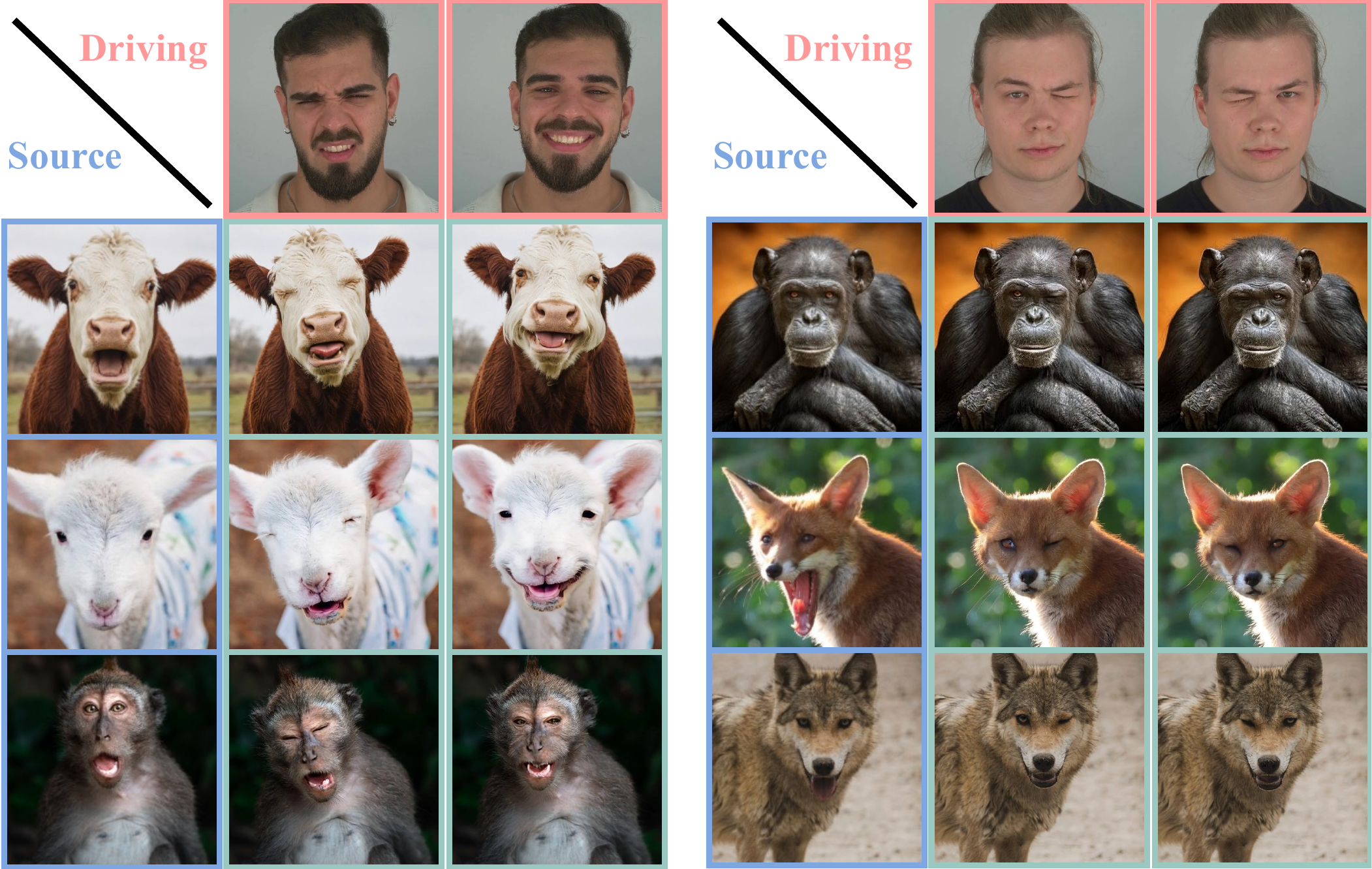}
    \caption{
Qualitative unseen-source generalization by \ours{}.
Our method transfers human driving motion while preserving the identity
and species-specific morphology of source portraits from unseen species.
\textcolor{src_blue}{\textbf{Blue}}, \textcolor{tgt_red}{\textbf{red}}, and
\textcolor{out_green}{\textbf{green}} borders denote
\textcolor{src_blue}{\textbf{source}} portraits,
\textcolor{tgt_red}{\textbf{driving}} images, and
\textcolor{out_green}{\textbf{CrossPortrait outputs}}, respectively.
The driving species is human and is covered by training; this figure
therefore evaluates source-side generalization only and does not evaluate
unseen-species driving.
    }
    \label{fig:unseen_source_species}
\end{figure}

\section{Data and Evaluation Protocol}
\label{app:data_protocol}

\subsection{Training Data}
\label{app:data_training}

Both stages are trained on the same underlying human--cat--dog collection,
which combines the human portrait datasets
NeRSemble~\citep{kirschstein2023nersemble},
AVA-256~\citep{martinez2024codec},
PFHQ~\citep{chen2023bfrffusion}, and
VFHQ~\citep{xie2022vfhq} with internally collected cat and dog portrait videos.
Each animal domain contributes approximately 15 hours of data.
For Stage 2, the same collection is organized into a merged metadata index,
where the known driving-species label $s_{\mathrm{drv}}$ selects the
corresponding ExpNet route.
Both stages follow the same portrait detection and cropping pipeline.
Stage 1 feeds $256\times256$ crops to PoseNet and ExpNet and supervises
reconstruction at $512\times512$, whereas Stage 2 trains on
$512\times512$ video clips with MotionNet inputs resized to
$256\times256$.

\subsection{Evaluation Splits}
\label{app:data_splits}

All source identities and driving clips used for evaluation are held out from
training.
The animal self-reenactment subset contains 50 cat and 50 dog videos collected
from Pixabay~\citep{pixabay} and Pexels~\citep{pexels}, each cropped to
$512\times512$ and trimmed to 3--5 seconds.
Beyond self-reenactment, the evaluation includes H$\to$A, A$\to$H,
cross-species A$\to$A, same-species animal cross-identity transfer, and a
qualitative unseen-source study with trained-species drivers
(Fig.~\ref{fig:unseen_source_species}).
\subsection{Pairwise User-Study Protocol}
\label{app:data_user_study}

For each pairwise user-study setting, driving clips are balanced across five
motion types: head rotation, head nodding, eye closure, mouth opening, and
mixed motion. This selection emphasizes motions with meaningful
correspondence across species and avoids over-weighting a single expression
or pose pattern. For H$\to$A, cat and dog source portraits are equally
represented so that the preference rates are not dominated by one animal
species.

Each comparison uses the same source--driving pair for the methods being
compared. Participants judge expression fidelity, head-pose agreement, source
identity and morphology preservation, and overall visual quality in separate
preference questions. A tie contributes half a vote to each method, so a
reported preference rate is computed as the wins plus half of the ties divided
by the number of valid comparisons. This protocol makes the preference
columns comparable across settings while keeping the motion distribution
balanced.

\end{document}